\documentclass{article}

\usepackage{arxiv}

\usepackage[utf8]{inputenc} 
\usepackage[T1]{fontenc}    
\usepackage{hyperref}       
\usepackage{url}            
\usepackage{booktabs}       
\usepackage{amsfonts}       
\usepackage{nicefrac}       
\usepackage{microtype}      
\usepackage{lipsum}
\usepackage{graphicx}

\usepackage{amsmath,amssymb,amsfonts}
\usepackage{graphics} 
\usepackage{epsfig} 
\usepackage{times} 
\usepackage{subfig}
\usepackage{tabularx}
\usepackage{booktabs}
\usepackage{multirow}

\usepackage{enumitem}

\DeclareMathOperator*{\argmax}{arg\,max}

\graphicspath{ {./images/} }

\title{RL-DWA Omnidirectional Motion Planning for Person Following in Domestic Assistance and Monitoring}

\author{
  Andrea Eirale, \\
  Department of Electronics\\ and Telecommunications (DET)\\
  Politecnico di Torino \\
  Torino, Italy \\
  \texttt{andrea.eirale@polito.it} 
   \And
  Mauro Martini \\
  Department of Electronics\\ and Telecommunications (DET)\\
  Politecnico di Torino \\
  Torino, Italy \\
  \texttt{mauro.martini@polito.it} \\
  \And
  Marcello Chiaberge \\
  Department of Electronics\\ and Telecommunications (DET)\\
  Politecnico di Torino \\
  Torino, Italy \\
  \texttt{marcello.chiaberge@polito.it} \\
}

\begin{document}
\maketitle
\begin{abstract}
Robot assistants are emerging as high-tech solutions to support people in everyday life. Following and assisting the user in the domestic environment requires flexible mobility to safely move in cluttered spaces.
We introduce a new approach to person following for assistance and monitoring. Our methodology exploits an omnidirectional robotic platform to detach the computation of linear and angular velocities and navigate within the domestic environment without losing track of the assisted person. While linear velocities are managed by a conventional Dynamic Window Approach (DWA) local planner, we trained a Deep Reinforcement Learning (DRL) agent to predict optimized angular velocities commands and maintain the orientation of the robot towards the user. We evaluate our navigation system on a real omnidirectional platform in various indoor scenarios, demonstrating the competitive advantage of our solution compared to a standard differential steering following.
\end{abstract}

\keywords{Person Following \and Robot Assistant \and Deep Reinforcement Learning \and Human-Centered Navigation}

\section{Introduction}
\label{sec:Intro}
In recent years, population ageing and pandemics have been demonstrated to cause isolation of older adults in their houses, generating the need for a reliable assistive figure. Service robotics recently emerged as high-tech support to the problem, providing a series of aid functionality to satisfy daily indoor assistance. Robotic solutions take care of interactive social aspects \cite{gongora2019social} or monitoring the health status of the user \cite{gasteiger2021friends, 8448739}.

\begin{figure}
    \centering
    \includegraphics[width=0.32\textwidth]{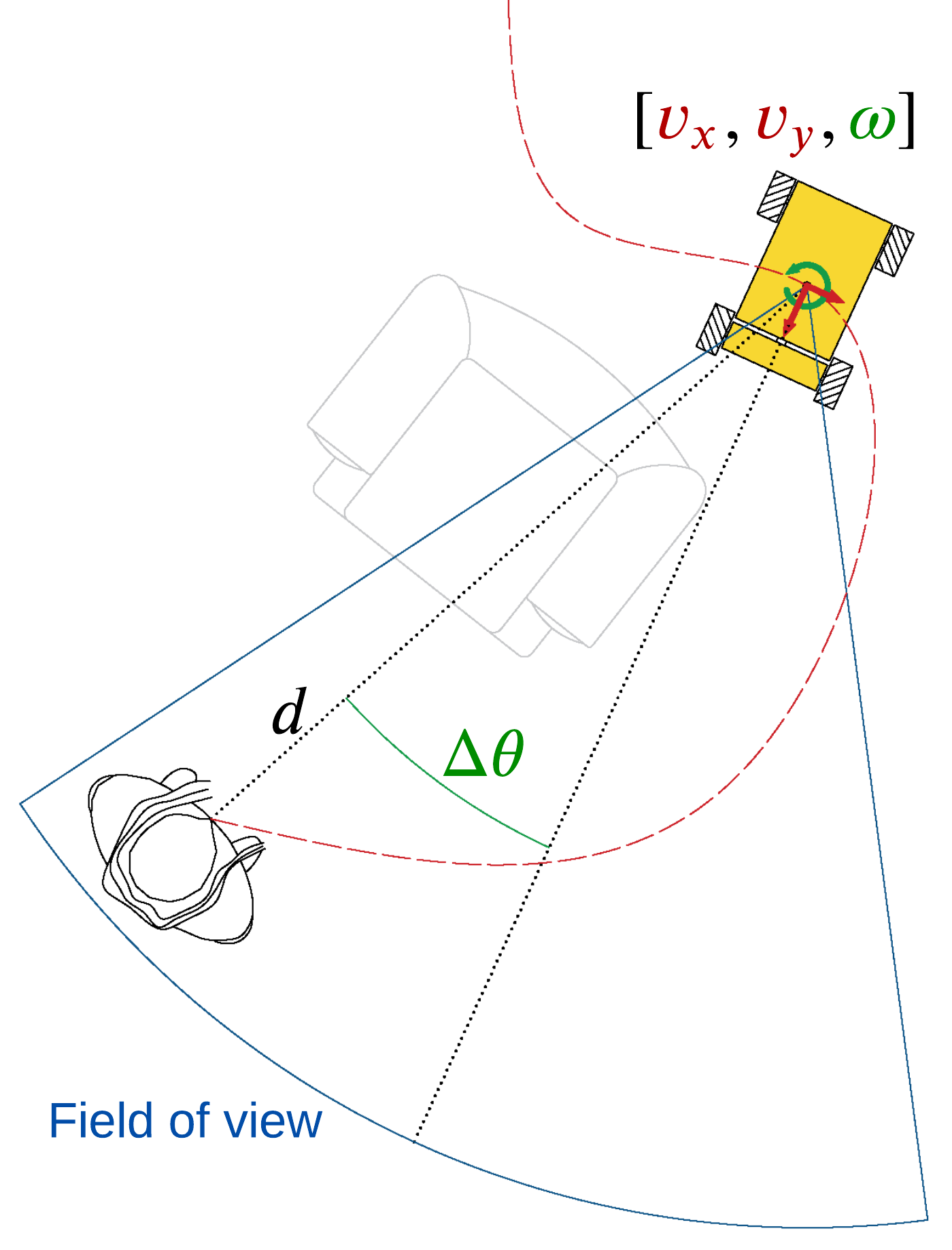} 
    \caption{Our person following solution exploits DWA planner to control linear velocities $[v_x, v_y]$ while the DRL agent controls the angular velocity $\omega$ to minimize the angle $\Delta\theta$ and maintain the user within the camera's field of view. This approach allows the rover to keep monitoring the person while avoiding obstacles on a safe trajectory.}
    \label{fig:nav_drawing}
\end{figure}

Domestic environments are often very demanding for autonomous navigation systems due to the variety of complex and dynamic obstacles they can feature. To this end, the robot platform shall provide extreme flexibility and effective mobility to handle narrow passages thought for humans. Moreover, in order to properly assist the user, the platform should be able to follow them within this environment.
Person following \cite{islam2019person, honig2018toward} is the first step to enable any visual or vocal interaction with the user while monitoring its condition to intervene earlier in the case of anomalous events. Person following systems are often based on naive visual-control strategy, directly coupling the generation of heuristic commands for the robot with the person coordinate in the image \cite{boschi2020cost}.
Deep Reinforcement Learning (DRL) agents recently demonstrated significant autonomy and flexibility boost in robotic solutions. Most works focus on training an end-to-end visual-control agent, which decides which velocity command actuates to follow the person recognized in the received color image (\cite{pang2020efficient, dat2021supporting}). 
Nonetheless, person following with omnidirectional mobile robot has been seldom investigated, and the attempts presented in literature are poorly correlated with rigorous methodologies and experimentations \cite{zhang2018autonomous,chen2017design}. This solution reduces the necessity for a dramatically precise visual tracking algorithm, and it results in being more cost-efficient compared to the adoption of an omnidirectional camera for a $360^\circ$ Field-Of-View (FOV), \cite{kobilarov2006people} or also a rotating camera such as gimbal systems, typical of Unmanned Aerial Vehicles (UAVs) following \cite{huh2013integrated}.

\subsection{Contribution}
In this work, we focus our research on developing a human-centered autonomous navigation system for a robotic assistant, which aims to fulfill the user monitoring requirement. We chose a tiny-size omnidirectional robotic base platform to fully exploit its kinematic advantages and propose an optimized person following methodology, always guaranteeing collision-free trajectory planning combined with continuous visual tracking of the user. We first set up a real-time perception pipeline based on PoseNet \cite{PoseNet} to identify the person and track their pose. Then, we train a Deep Reinforcement Learning (DRL) agent in a realistic 3D simulated environment, presented in \cite{martini2022pic4rl}, to follow a dynamic goal. The agent learns to optimize the yaw angular velocity $\omega$ to maintain the rover's orientation towards the goal (Figure \ref{fig:nav_drawing}). We integrated the DRL agent with a Dynamic Window Approach (DWA) local planner, which separately provides linear velocities $(v_x, v_y)$ to follow the person on a safe trajectory.

The contributions of this work are manifold: 
\begin{itemize}
    \item We study novel advantages of adopting an omnidirectional robot assistant for the person following task
    
    \item We design and train a Deep Reinforcement Learning agent to constantly re-orient the omnidirectional platform towards the moving person
    
    \item We effectively integrate the DRL agent with a navigation algorithmic stack to separately handle trajectory generation for obstacle avoidance and orientation control for person monitoring
    
    \item We set up a simple real-time perception pipeline to extract the coordinate of the person and visually track its pose
\end{itemize}

Nonetheless, compared to most previous works, we carried out extensive experimentations with the robot. To this end, we set up an innovative experimental framework based on an Ultra-Wideband (UWB) anchors system to localize both the person and the robot while moving and measure their relative distance and orientation. Our results validate the performance of our solution and show the competitive advantage and robustness it can provide in visually monitoring the user while avoiding obstacles in a cluttered indoor environment such as a domestic one.

\section{Methodology}
\label{sec:methodoogy}
In order to efficiently follow and monitor the user, the autonomous platform should always be aware of the subject's position during its navigation, which means keeping its orientation towards the person and maintaining them in the camera's field of view. This task might raise some serious difficulties in the case of conventional differential drive platforms, which do not have the possibility to describe a curved motion without a change in orientation. This limitation often leads differential drive robots to lose the human target while avoiding obstacles or following an occluded path. Therefore, it is clear that maintenance of a certain desired orientation and collision-free navigation towards a precise destination results in conflicting objectives.
On this basis, we propose a novel system to handle person following and monitoring in cluttered and unstructured environments, using an omnidirectional robotic platform.

\subsection{Perception and Tracking}
\label{sec:perception}
An essential requirement to visually monitor a person within the environment is the development of a real-time detection and tracking system. 
In this work, we developed a deep learning perception pipeline that allows the robot to track the person visually.
A RealSense D435i Camera, mounted on the rover at a human height, is used to collect color and depth images of the environment. In a first step, the presence of the person is detected through PoseNet \cite{PoseNet}, a lightweight deep neural network that estimates the pose of humans in images and videos. For each person in the scene, the network outputs the position of 17 key joints (like elbows, shoulders, or feet). In our implementation, PoseNet runs on the Google Coral Edge TPU device\footnote{https://coral.ai} at 30 frame-per-second (FPS), which corresponds to the maximum frame rate supported by the RealSense D435i camera. The key points predicted by PoseNet are then translated into a bounding box that localizes the person within the image. The resulting bounding box is tracked with SORT \cite{SORT}. The central point $C$ is computed as the average of shoulders or hip joints. This structure guarantees reliable esteem of the person's position in the environment to be fully usable by the robot navigation system, avoiding the risk of inaccurate motion planning. The distance of the person from the robot $d_C$ is then extracted from the depth frame as the value corresponding to the point $C$. The complete information contained in the resulting array $(x_C,y_C, d_C)$ can be easily translated into the person position within the environment, $(x_P,y_P)$, with basic reference frame transformations.

\subsection{Deep Reinforcement Learning Agent for Person-focused Orientation Control}
\label{sec:drl_agent}
As introduced before, we model the angular velocity control according to a reinforcement learning framework. Therefore, the problem is formulated as a Markov Decision Process (MDP) described by the tuple $(\mathcal{S},\mathcal{A}, \mathcal{P}, R, \gamma)$ \cite{sutton2018reinforcement}. An agent starts its interaction with the environment in an initial state $s_0$, drawn from a pre-fixed distribution $p(s_0)$ and then cyclically select an action $\mathbf{a_t} \in \mathcal{A}$ from a generic state $\mathbf{s_t} \in \mathcal{S}$ to move into a new state $\mathbf{s_{t+1}}$ with the transition probability $\mathcal{P(\mathbf{s_{t+1}}|\mathbf{s_t},\mathbf{a_t})}$, receiving a reward $r_t = R(\mathbf{s_t},\mathbf{a_t})$.

A reinforcement learning process aims to optimize a parametric policy $\pi_\theta$, which defines the agent behavior once trained. In the context of autonomous navigation, we model the MDP with an episodic structure with maximum time steps $T$. Hence, the agent is trained to maximize the cumulative expected reward $\mathbb{E}_{\tau\sim\pi} \sum_{t=0}^{T} \gamma^t r_t$ over each episode, where $\gamma \in [0,1)$ is the discount factor. More in detail, we use a stochastic agent policy in an entropy-regularized reinforcement learning setting, in which the optimal policy $\pi^*_\theta$ with parameters $\mathbf{\theta}$ is obtained maximizing a modified discounted term:
\begin{equation}
    \pi^*_\theta = \argmax_{\pi} \mathbb{E}_{\tau\sim\pi} \displaystyle \sum_{t=0}^{T} \gamma^t [r_t + \alpha \mathcal{H}(\pi(\cdot|s_t))]
\end{equation}
Where $\mathcal{H}(\pi(\cdot|s_t))$ is the entropy term that increases robustness to noise through exploration, and $\alpha$ is the temperature parameter that regulates the trade-off between reward optimization and policy stochasticity. 

We train the agent's policy neural network with the Soft Actor-Critic (SAC) algorithm presented in  \cite{haarnoja2018soft}, allowing a continuous action space. In particular, we instantiate a stochastic Gaussian policy for the actor and two Q-networks for the critics.

\begin{description}[leftmargin=0pt]
    \item[Input features] The input features of the policy network embed the necessary information about the dynamic goal:
    
    1) $d_t$: the distance of the goal from the rover
    
    2) $\Delta\theta_t$: the angular difference between the orientation of the rover and the orientation of the vector connecting the rover's center of rotation with the goal (Figure \ref{fig:nav_drawing})
    
    3) $\omega_{t-1}$: yaw velocity command assigned to the platform at the previous time instant
    
    \item[Reward] Reward shaping is the typical process that leads researchers to analytically specify the desired behavior to the agent thanks to a dense reward signal assigned at each time step. 
    To this end, we define a reward $r_h$ as the arithmetic sum of two distinct contributions:
    \begin{gather}
        r_{yaw} = \left( 1 - 2 \sqrt{\left| \frac{\Delta\theta_t}{\pi} \right|} \right)\\
        r_{smooth} = -|\omega_{t-1} - \omega_t|
    \end{gather}
    The first contribute $r_{yaw}$ teaches the agent to maintain its orientation towards the goal, while the second contribute $r_{smooth}$ is used to obtain a smooth transition between the current agent's yaw velocity output and that at the next time instant.
    
    \item[Action] The DRL agent computes the angular velocity $\omega$ at any time instant. The two velocities provided by the DWA planner are merged in a unique velocity command $V = [v_x, v_y, \omega]_t$, and executed by the robotic platform.
    
    \item[Neural network architecture] The simple neural network used for the orientation control policy comprises three dense layers, respectively with $512$, $256$, and $256$ units each.
\end{description}

\subsection{Omnidirectional Motion Planner and Obstacle Avoidance}
\label{sec:Navigation_planner}
We develop an autonomous navigation solution that handles the generation of collision-free trajectories and the control of the platform orientation separately. This allows the rover to reach various destinations in a domestic, cluttered space while continuously monitoring the subject of interest.

In order to compute an obstacle-free trajectory towards a goal, the system needs to acquire the rover's pose (position and orientation) with respect to a fixed reference frame and perceive obstacles around it. In our implementation, we exploited a RealSense T265 Tracking Camera to obtain information about the rover's pose and an RPLiDAR A1 LiDAR to retrieve 2D laserscan distance measurements of the obstacles around the robot.

We developed our navigation system tailoring the Navigation2 navigation stack\footnote{https://navigation.ros.org/} for the specific use case of person following and monitoring. 

The resulting navigation system consists of a DWA local planner and controller, which receives the coordinate of the navigation goal within the environment $(x_G, y_G)$ (corresponding to the person's position), and computes the linear velocity commands $[v_x, v_y]$ for the omnidirectional platform to follow the predicted collision-free trajectory.
At any time instant, these velocities are merged with the angular yaw velocity provided by the DRL agent, and the whole velocity command $[v_x, v_y, \omega]$ is passed and executed by the robotic platform. 
Moreover, a plugin specifically developed for this study is added to truncate the navigation path shortly before the goal. This ensures a secure distance from the user at any time.

\begin{figure}[h]
    \centering
	    \includegraphics[width=0.4\textwidth]{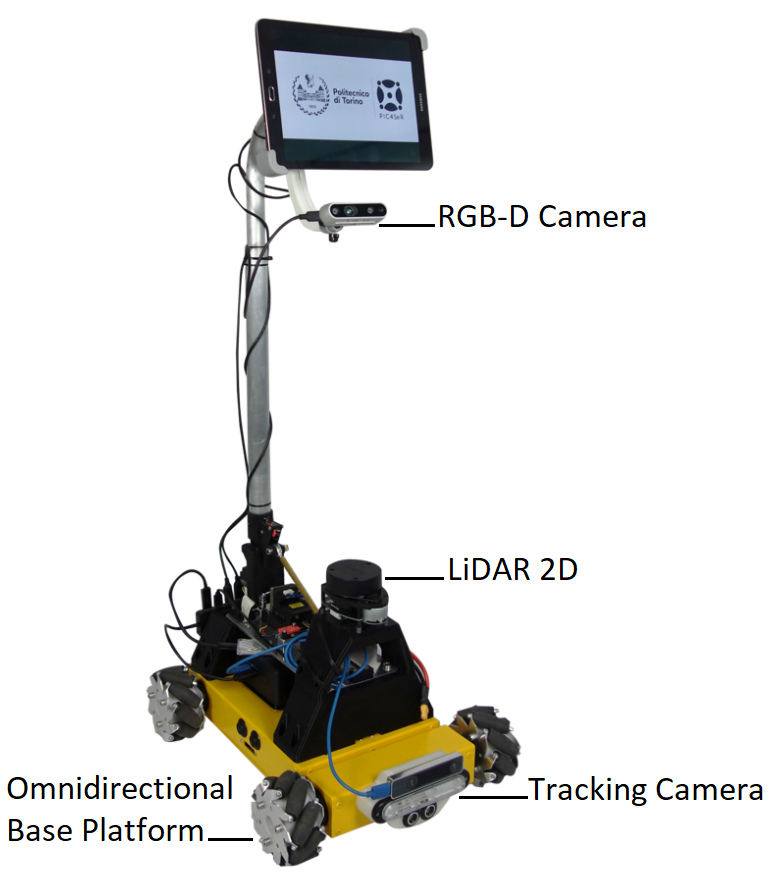}
	    \caption{The omnidirectional platform we set up for experimentation and validation of our novel methodology.}\label{fig1}
	    \label{fig:Marvin}
\end{figure}

\section{Experiments and Results}
\label{sec:experiments}

\subsection{Agent Training Setting}
Our model is obtained from the customization of a SAC agent from the TF2RL library \footnote{https://github.com/keiohta/tf2rl}. Actor and critic networks present respectively $218,114$ and $218,369$ parameters and are trained with Adam optimizer and a learning rate of $2\cdot10^{-4}$. The $\epsilon$-greedy exploration policy is defined by the starting value $\epsilon_0 = 1.0$, the decay $\gamma_\epsilon=0.992$ and a minimum value for random action sampling of $\epsilon_{min} = 0.05$. The agent is trained for $3300$ episodes composed of $T=300$ maximum steps. The robot's starting pose is changed every $20$ episodes to guarantee a good level of exploration and resulting generalization.

In order to merge all the software components and technologies needed to perceive and navigate the environment, we decided to adopt the PIC4rl-gym, \cite{martini2022pic4rl}, which relies on the Robot Operating System 2 (ROS2) Foxy and the open-source simulator Gazebo for training the agent.
The simulation environment is composed of a domestic scenario with several spaces and obstacles.
A Gazebo plugin simulates the dynamic goal the agent has to follow.

\subsection{Navigation Settings}
For our experimentation, we used the Marvin omnidirectional robotic platform shown in Figure \ref{fig:Marvin}, described in detail in \cite{eirale2022marvin}. Moreover, the robot features a vertical controllable shaft useful to raise the RGB RealSense camera over obstacles height. Its small footprint and the mobility given by the configuration of its wheels are optimal for navigating in cluttered, narrow environments.
\begin{table}[ht]
    \centering
    \caption{Results obtained from the person following test in four different scenarios. Our omnidirectional planning and control system clearly demonstrates a performance gap in keeping the tracking of the person while following its motion: the $\Delta\theta$ error is drastically reduced in comparison with a differential drive navigation.}
    \renewcommand{\arraystretch}{1.5}
    \begin{tabularx}{0.7\textwidth} { 
        >{\raggedright\arraybackslash}X 
        >{\centering\arraybackslash}X 
        >{\centering\arraybackslash}X 
        >{\centering\arraybackslash}X 
        >{\centering\arraybackslash}X 
        >{\centering\arraybackslash}X }
        \hline
        \textbf{Scenario} & $\Delta\theta$ & \textbf{Mean} & \textbf{Std.Dev.} & {\textbf{RMSE}} & {\textbf{MAE}}\\
        \hline
        \textbf{1} & 
        Omnidir. & $3.36$ & $10.45$ & $10.71$ & $8.57$\\
         & Differential & $16.00$ & $63.41$ & $68.31$ & $57.20$\\
        \hline
        \textbf{2} & 
        Omnidir. & $-4.35$ & $8.71$ & $9.45$ & $8.08$\\
         & Differential & $-15.67$ & $53.99$ & $58.48$ & $50.11$\\
        \hline
        \textbf{3} &
        Omnidir. & $0.35$ & $8.26$ & $8.33$ & $6.61$\\
         & Differential & $12.34$ & $42.19$ & $45.05$ & $37.38$\\
        \hline
        \textbf{4} & 
        Omnidir. & $4.56$ & $11.28$ & $13.13$ & $10.58$\\
         & Differential & $27.66$ & $20.95$ & $35.07$ & $29.19$\\
        \hline
    \end{tabularx}
    \label{tab:FollowingResults}
\end{table}

\subsection{Results}
\begin{figure*}[ht]
    \centering
    
    \subfloat[][\emph{Scenario 1 - Omnidirectional configuration}]
        {\includegraphics[width=35mm]{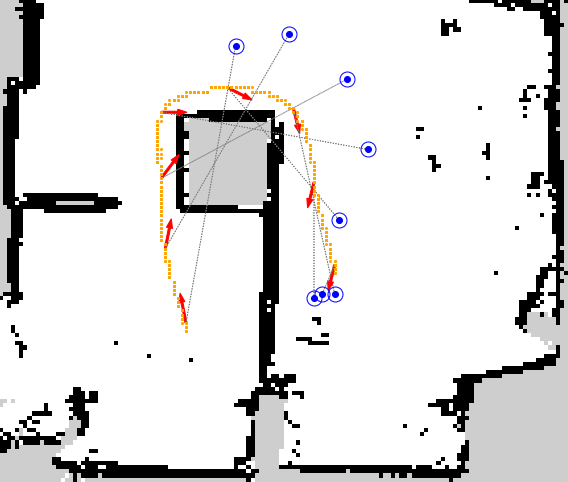}} \quad
    \subfloat[][\emph{Scenario 1 - Differential configuration}]
        {\includegraphics[width=35mm]{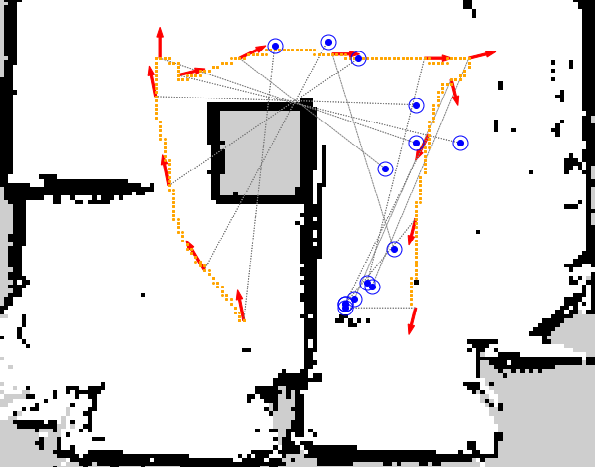}} \quad
    \subfloat[][\emph{Scenario 2 - Omnidirectional configuration}]
        {\includegraphics[width=35mm]{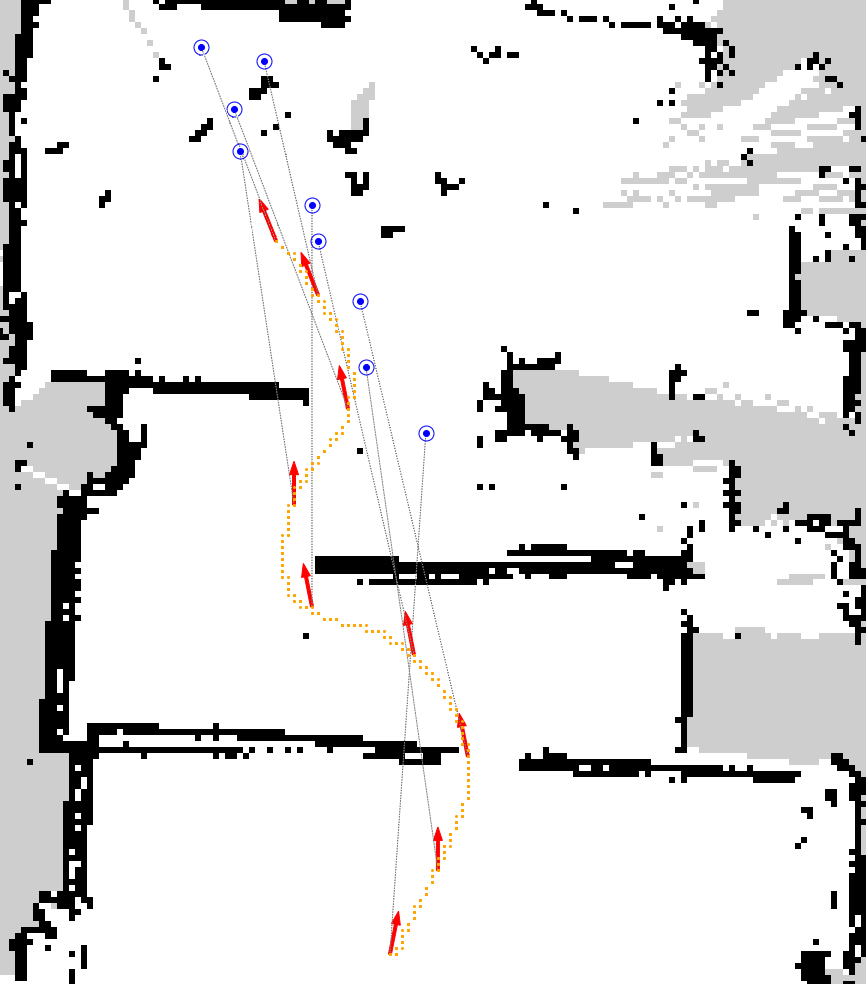}} \quad
    \subfloat[][\emph{Scenario 2 - Differential configuration}]
        {\includegraphics[width=35mm]{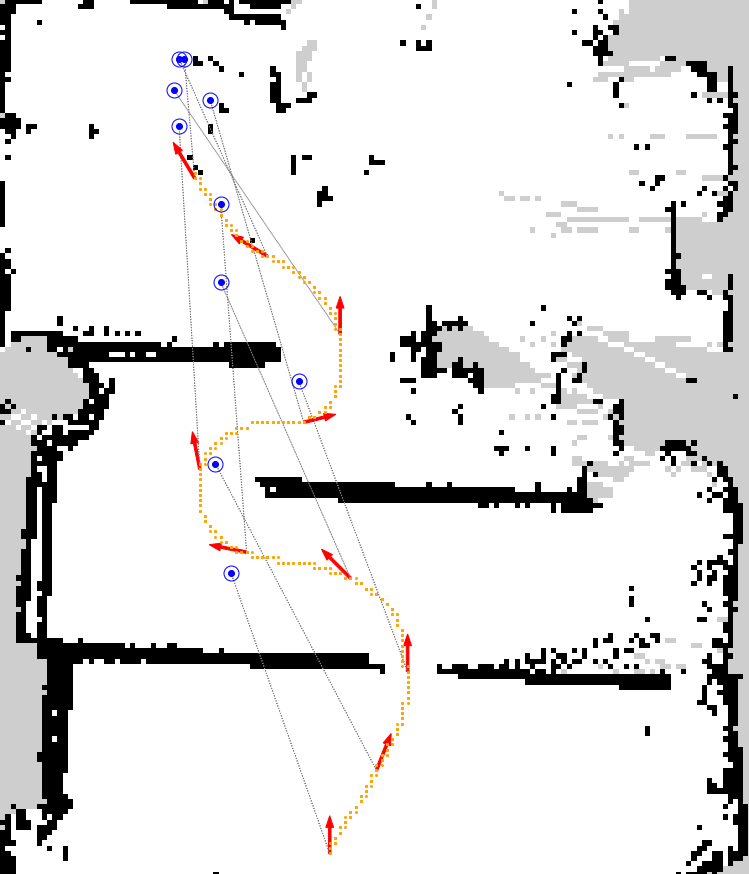}} \\
    \subfloat[][\emph{Scenario 3 - Omnidirectional configuration}]
        {\includegraphics[width=35mm]{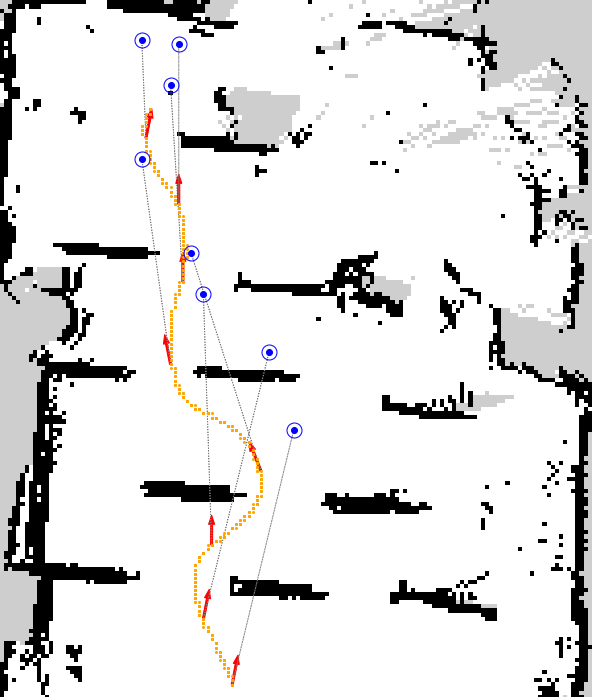}} \quad
    \subfloat[][\emph{Scenario 3 - Differential configuration}]
        {\includegraphics[width=35mm]{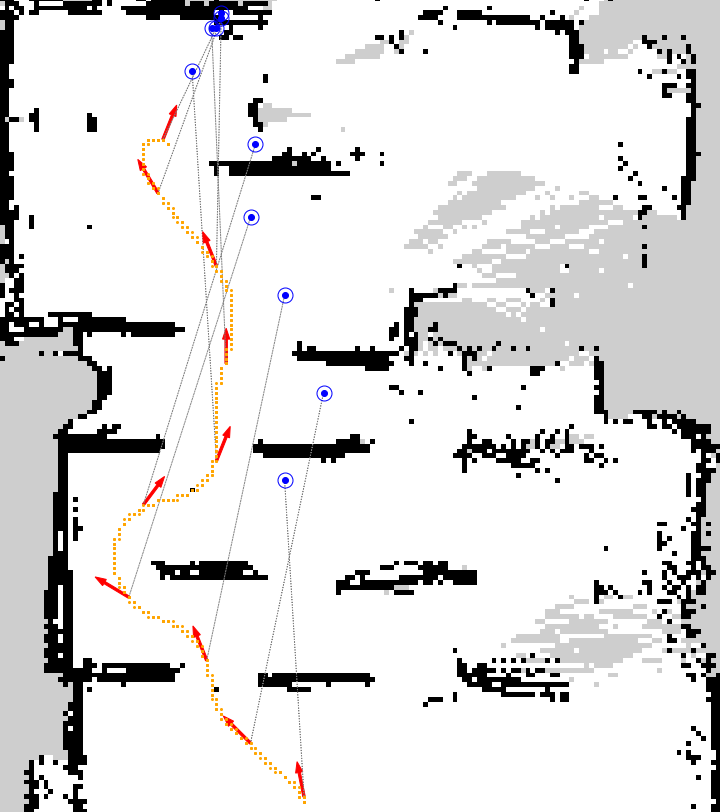}} \quad
    \subfloat[][\emph{Scenario 4 - Omnidirectional configuration}]
        {\includegraphics[width=35mm]{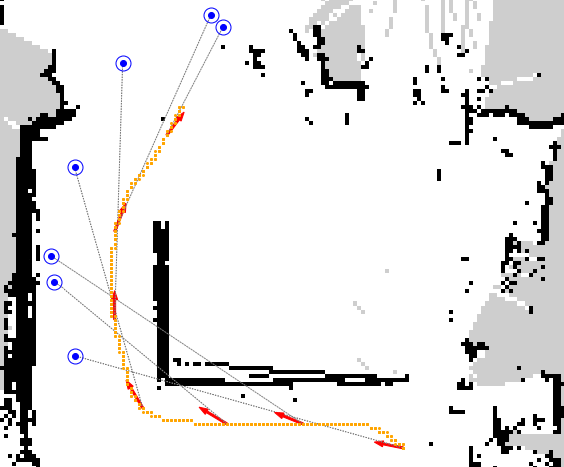}} \quad
    \subfloat[][\emph{Scenario 4 - Differential configuration}]
        {\includegraphics[width=35mm]{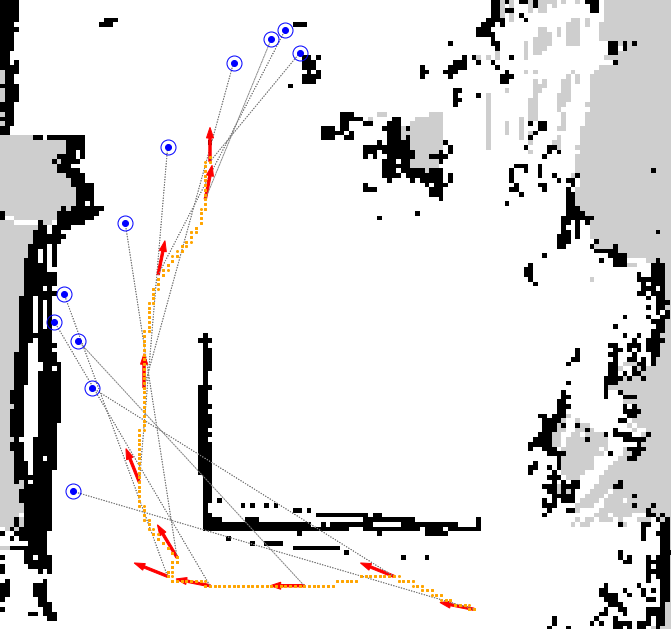}}
        
    \caption{Person following results: scenario 1 is composed of a wide U-shaped path, scenario 2 presents narrow passages through obstacles, scenario 3 presents a high number of obstacles and possible paths, while scenario 4 is composed of a high $90^\circ$ wall to be circumnavigated. Red arrows indicate position and orientation of the rover associated with the person's position (blue point) at the same instant. The orange spline represents the path crossed by the rover.}
    \label{fig:following_results_A}
\end{figure*}

Tests are performed in four different scenarios. The person moves for the whole extent of the test, and the rover has to follow them, using the position $(x_P, y_P)$ extracted from the visual perception pipeline as a dynamic goal of the navigation. For this reason, to ensure an accurate ground truth data collection, we set up a localization system based on four ultra-wideband anchors placed in the testing area. One additional anchor is placed upon the rover, and the followed person holds a second one. The rover's orientation is also aligned with the one used by the ultra-wideband system. In such a way, it is possible to know the actual relative position between the rover and the followed person. This allows us to correctly compute the angular difference $\Delta\theta$ at any time instant. To our knowledge, this experimental setting is the first attempt in the literature to quantitatively measure the quality of a person following system performance, going beyond the typical qualitative evaluation. Our extensive experimentation with the ultra-wideband localization system provides significant proof of the robustness of our navigation solution for person following.

For each scenario, tests are performed with the robot in Figure \ref{fig:Marvin} comparing a standard differential drive configuration, hence without using velocity $v_y$, with our RL-DWA novel navigation methodology.

Seven validation runs are performed for every rover configuration and scenario. Considered metrics for each test are the average error $\Delta\theta$ with its standard deviation, the root mean square error (RMSE), and the mean absolute error (MAE) maintained along the whole path. Table \ref{tab:FollowingResults} reports the average value computed for each scenario and metric over all the different tests.

Furthermore, in Figure \ref{fig:following_results_A}, for each scenario and configuration, a visualization of the performed test is reported. The gridmaps reported in the figure are directly obtained from the rover during the navigation, while rover and person poses are obtained from the ultra-wideband system.
As can be seen, our methodology proves to robustly track the followed person more effectively than a traditional differential drive navigation in all the considered scenarios. In the omnidirectional configuration (Figure \ref{fig:following_results_A}a, \ref{fig:following_results_A}c, \ref{fig:following_results_A}e, \ref{fig:following_results_A}g) the rover manages to always maintain the user within the camera's view, contrary to the differential drive case, where the visual contact is instead lost several times. This generally leads to higher performance in following the user, with the rover planning more optimal collision-free trajectories, fully satisfying also the person monitoring requirement. The obtained values of $\Delta\theta$ clearly show the performance gap in all scenarios, demonstrating the successful behavior in monitoring the person provided by our solution. Also in the fourth scenario (Figure \ref{fig:following_results_A}g, \ref{fig:following_results_A}h), where after the curve the wall obstructs the rover's view of the user, it appears clear that the ability to remain facing the dynamic human goal is beneficial for a more accurate re-acquisition of tracking as soon as the obstacle is passed.

\section{Conclusions}\label{sec:conclusion}
In this work, we propose a novel, cost-effective approach for autonomous person following in the context of domestic robotic assistance. Differently from previous works, we aim at keeping the robot oriented toward the human during the whole navigation avoiding costly sensors for continuous tracking.
We first set up a real-time visual perception pipeline to identify the person and estimate their pose. Then, we train a DRL agent in a realistic 3D simulated environment to compute optimized yaw angular velocity $\omega$ to maintain the rover’s orientation towards the person. We merge this velocity contribution with a DWA omnidirectional controller and propose a system that treats orientation control and dynamic trajectory planning separately. This allows the system to fulfill both the monitoring and the obstacle avoidance objectives of the robotic assistive task.
We conduct extensive experimentation, adopting an Ultra-Wide Band localization system as ground truth, and demonstrate how our solution outclasses conventional differential drive platforms.
To our knowledge, this is the first study combining omnidirectional motion planning and DRL to enable a robust person following.
\bibliographystyle{unsrt}  
\bibliography{biblio}  

\end{document}